# ECG Arrhythmia Detection Using Disease-specific Attention-based Deep Learning Model

Linpeng Jin

Hangzhou Normal University, Hangzhou, 311121, China

**Abstract:** The electrocardiogram (ECG) is one of the most commonly-used tools to diagnose cardiovascular disease in clinical practice. Although deep learning models have achieved very impressive success in the field of automatic ECG analysis, they often lack model interpretability that is significantly important in the healthcare applications. To this end, many schemes such as general-purpose attention mechanism, Grad-CAM technique and ECG knowledge graph were proposed to be integrated with deep learning models. However, they either result in decreased classification performance or do not consist with the one in cardiologists' mind when interpreting ECG. In this study, we propose a novel disease-specific attention-based deep learning model (DANet) for arrhythmia detection from short ECG recordings. The novel idea is to introduce a soft-coding or hard-coding waveform enhanced module into existing deep neural networks, which amends original ECG signals with the guidance of the rule for diagnosis of a given disease type before being fed into the classification module. For the soft-coding DANet, we also develop a learning framework combining self-supervised pre-training with two-stage supervised training. To verify the effectiveness of our proposed DANet, we applied it to the problem of atrial premature contraction detection and the experimental results shows that it demonstrates superior performance compared to the benchmark model. Moreover, it also provides the waveform regions that deserve special attention in the model's decision-making process, allowing it to be a medical diagnostic assistant for physicians.

**Key Words:** ECG, Deep Learning, Attention mechanism, Self-supervised Pre-training

## 1. Introduction

In recent decades, the prevalence of cardiovascular diseases is continuously increasing and will still remain an upward trend in China [1]. As a noninvasive and inexpensive tool, the standard 12-lead electrocardiogram (ECG) is one of the most commonly-used measures to diagnose cardiovascular diseases. With the rapid development of artificial intelligence and the wide application of communication technology, the computer-aided ECG diagnostic tool plays an increasingly important role in preventing cardiovascular diseases and saving patients' lives.

The traditional methods for arrhythmia classification heavily rely on discriminative features derived from ECG signals and the performance is unsatisfactory due to the unbreakable bottleneck in feature extraction. In recent years, deep learning represented by convolutional neural network (CNN) [2] and long short-term memory network (LSTM) [3] has made major breakthroughs in the field of ECG analysis and greatly improve classification performance. However, we cannot get any other valuable reference information (such as the underlying logic for interpreting ECGs and lesion locations) except the diagnosis conclusion, which seriously restricts the application and popularization of deep-learning models in clinical practice. In addition, the performance for detecting different arrhythmia types from short ECG recordings may vary widely. For example, the detection accuracy of atrial premature contraction (APC) is much lower than that of atrial fibrillation via deep-learning models with the same network architecture. The root cause for this phenomenon is that deep neural networks treat each sampling point of an ECG signal equally. As we can see from Fig.1, P wave and T wave have much lower amplitudes and shorter durations than QRS complex. Some abnormal ECG signals even have no P waves or only have flat T waves, which further strengthen the dominance of QRS complexes. The common practice is to feed ECG signals into deep neural networks (DNN) directly. Due to the above-mentioned properties of ECG waveforms, it will result in an undesirable and unwanted consequence: deep neural networks only effectively captured discriminative features derived from QRS complexes, rather than the ones derived from P and T waves. Because of this, they achieve inferior performance when used to detect cardiac arrhythmia that is mainly related to the features in non-QRS complexes.

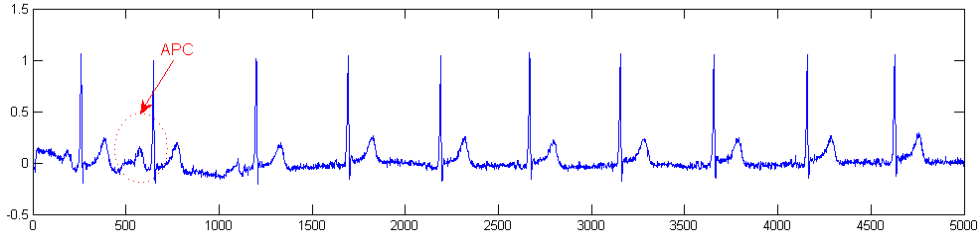

Fig.1 A raw ECG signal from a patient with APC

In clinical practice, physicians make a diagnosis according to the morphologies of P waves, QRS complexes and T waves in ECG signals, and the durations between them. The rules for diagnosis of different arrhythmia types vary widely, thus different waveform regions of an ECG signal should be treated differently. For example, we need to pay closer attention to P wave for APC, and to the regions between S wave and T wave for ST-T abnormalities. This idea is somewhat like attention mechanisms that have been widely used in natural language processing [4], computer vision [5] and time series analysis [6-7]. However, the attention mechanism that we are familiar with is general-purpose with no need for domain knowledge. It often serves as a network component and is integrated into an end-to-end deep learning framework. Such a mechanism can effectively improve performance, but the attention weights obtained in a data-driven way do not always consist with the ones in cardiologists' mind when they interpret ECG [8]. In the worst case, attention weights become meaningless and even provide incorrect and misleading information to primary physicians.

To overcome the above discussed problem, we propose a disease-specific attention-based deep learning model (DANet) for ECG analysis, which amends original ECG signals via cardiological domain knowledge before feeding them into deep neural networks. Specifically, we enable the DANet to have the ability of mimics cardiologists' cognitive processes in diagnostic decision making by introducing a disease-specific attention mechanism. The rest of this paper is organized as follows: in section 2, we introduce existing studies that have similar design goal with our solution and section 3 presents a detailed description of the proposed DANet; a case study as well as experimental results are shown in section 4; discussion and conclusions are provide in section 5 and 6 respectively.

## 2. Related Works

To highlight the novelty of the proposed DANet, in what follows we will briefly discuss existing deep-learning models related to our work.

In consideration of the fact that different channels and temporal segments of a feature map extracted from the ECG signal contribute differently to arrhythmia detection, Zhang [9] proposed spatio-temporal attention-based convolutional recurrent neural network to focus on representative features along both spatial and temporal axes. It is a typical application example of general-purpose attention mechanisms, which only utilizes the data-driven technology. Jo [10] developed an explainable deep learning model based on a neural network-backed ensemble tree with six feature modules that are able to explain the reasons for its decisions. The key insight is that this model uses decision trees for high-level interpretability and employs neural networks for low-level decisions. Neves [11] adopted three model-agnostic methods including Permutation Feature Importance, Local Interpretable Model-agnostic explanations and Shapley additive explanations to make heartbeat classification more explainable. Lu [12] added the rhythm variation coefficient to CNNs as a symbolic representation of time series for the sake of performance improvement and clinical usability. Le [13] introduced a lead-wise attention module to aggregate outputs from three one-dimensional convolutional neural networks and employed the Grad-CAM technique to produce a meaningful lead-wise explanation. Sun [14] devised a knowledge inference module containing an ECG knowledge base and a rule-grounding, matching and scoring module. It is trained together with a baseline DNN classifier by the joint learning module using the back-propagation algorithm. Jin [15] developed a cardiologist-level interpretable knowledge-fused deep neural network for ECG arrhythmia detection, in which a deep learning module including CNN feature extractor, spatial attention, bidirectional LSTM and multi-label classifier outputs initial results, and a medical post-processing module using R peak-based rule inference obtains the final results. Ge [16] designed an ECG knowledge graph framework for abnormal electrocardiogram diagnosis. It utilizes ECG key point-based attribute features, knowledge graph embedding, graph convolutional network and expert knowledge to make a prediction.

There are also many research works that employs general-purpose attention mechanism, Grad-CAM technique, Shapley and knowledge graph to make deep learning models show interpretability results [6-7,17-21], but their technological routes are markedly different from ours. Tao [22] proposed an expert-knowledge attention network to detect tachyarrhythmia from ECG signals, which integrates a six-layer CNN and a gated recurrent unit (GRU) with lead and rhythm attention modules. The novelty is that rhythm attention score is calculated by a formula based on R-R intervals and lead importance score is provided based on the experience of clinical experts. That is to say, the two attention modules do not need any training. Although the idea that utilizes the knowledge of cardiology to construct attention modules is similar to ours, it is implemented in a different way.

## 3. Model Description

### 3.1 Overall Framework

The proposed DANet is mainly used for detecting one type of arrhythmia from ECG signals, namely binary classification problem. It is feasible to detect multiple arrhythmia types, but there will not be impressive interpretability results. Fig.2 shows the overall framework of the proposed model. As we can see, it is mainly composed of five processing modules: data preprocessing, manual attention weight, data augmentation, waveform enhanced module (associated with automatic attention weight) and classification module. In what follows, we will describe the technical details of each module.

(1) *Data Preprocessing*: It is very necessary to preprocess ECG data before feeding it into the subsequent deep-learning modules. The popular methods include resampling, filtering, amplitude normalization, lead selection and so on. We can adopt one or more data preprocessing techniques as needed. Meanwhile, a traditional fiducial point method such as ECGPuWave [23] is employed to find the onset and offset of every P wave, every QRS complex and every T wave in the input ECG recording.

(2) *Manual Attention Weight*: According to the clinical diagnostic rule of a given disease type and the onset and offset of each of the P wave, QRS complex and T wave obtained by the traditional fiducial points method, we manually set the attention weights for different waveform regions. To be specific, each sampling point of an ECG signal is assigned an attention weight, but the value for sampling points at the same position in different leads is the same. Note that there is no need to establish manual attention weights for P wave, QRS complex and T wave in the testing phase.

(3) *Data Augmentation*: It is recommended to employ data augmentation technology to train deep learning models since they can improve the generalization ability very efficiently. The commonly used strategies such as "picking random ECG segment" and "adding random noise" [24] can be used to deal with the preprocessed ECG data along with manual attention weights, so that we can get a rich training dataset.

(4) *Waveform Enhanced Module*: The feature extraction-oriented deep learning model plays the role of the waveform enhanced module. The prevalent network architectures such as dilated convolutional neural network [25] and U-Net [26] can be used for this purpose. The waveform enhanced module takes ECG data $\{d_{ik} | 1 \leq i \leq channelC, 1 \leq k \leq frameC\}$ as an input and outputs automatic attention weights $\{w^2_k | 1 \leq k \leq frameC\}$ through calculating each internal node state in forward propagation pass. Here *channelC* and *frameC* denote the number of ECG leads and sampling points respectively. This module needs to be trained in advance, and we can use the loss function concerning manual attention weights $\{w^1_k | 1 \leq k \leq frameC\}$ and automatic attention weights $\{w^2_k | 1 \leq k$

≤ $frameC$} to perform an end-to-end learning task. Since we can easily obtain $w^1_k$ via traditional fiducial point methods with no need to engage physicians to do such work, this training stage is also called self-supervised pre-training.

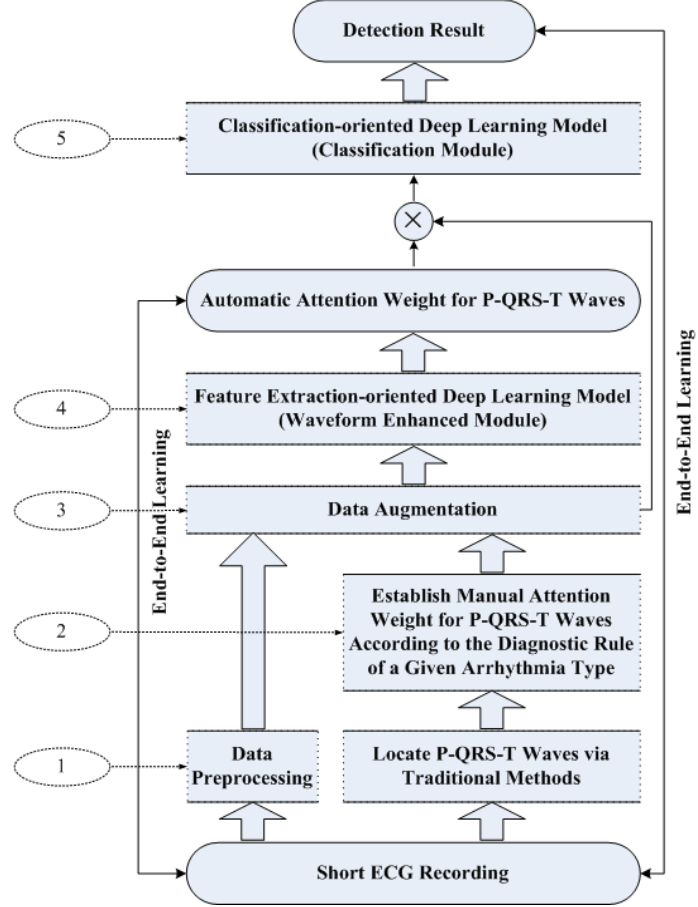

Fig.2 The overall framework of DANet

(5) *Classification Module*: The classification-oriented deep learning model is used to output the final detection result. Existing network models such as VGGNet [27] and SENet [28] are all qualified for the task. Unlike regular practices, ECG data {$d_{ik}$ |1≤ $i$ ≤ $channelC$, 1≤ $k$ ≤ $frameC$} is first multiplied by automatic attention weights {$w^2_k$ | 1≤ $k$ ≤ $frameC$} by sampling-point-wise operation, the resulting ECG data that imply cardiological domain knowledge is then fed into the classification module to calculate class probabilities. To prevent model overfitting, we propose a two-stage training method for it. In the 1st stage, we freeze all the parameters of the waveform enhanced module and train the classification module in an end-to-end manner using the back propagation algorithm. Due to the inherent defect of traditional fiducial point methods, manual attention weights may not be right and thus we should take remedial measures to fix this problem to some extent. That is the reason why the 2nd training stage is introduced to further fine-tune the whole model (including the waveform enhanced and classification modules) in an end-to-end fashion. Note that overfitting will occur if we train the whole model for too long. In the testing stage, we can get some valuable reference information from automatic attention weights besides detection results. For example, if automatic attention weights do not focus on the region between the S wave and T wave, it is meaningless when the detection result is ST-T abnormalities.

## 3.2 Soft-coding Attention of ECG waveforms

The core idea of the proposed DANet is that routine deep-learning models should utilize cardiological domain knowledge to treat different ECG wave regions differently, especially P wave, QRS complex and T wave with clinical diagnostic significance. In our model, the waveform enhanced module used to discover specific wave regions and the classification module used to predict detection results are designed to be compatible with many existing deep learning models. Fig.3 shows the network architecture of a specific DANet that has been proved to be effective in practice, in which a dilated convolutional neural network with residual connection and a classic convolutional neural network take on the roles of the waveform enhanced and classification modules respectively. The residual connection is helpful for gradient back-propagation, especially when we train very deep neural networks. The dilated convolution can effectively enlarge the receptive field without increasing computational burdens, which is beneficial for accurately locating P wave, QRS complex and T wave. To deal with variable-length ECG data, recurrent layers such as LSTM and GRU [29] can be used together with the classic CNN. An alternative solution is to adopt local pattern aggregation-based deep-learning models as the classification module [30].

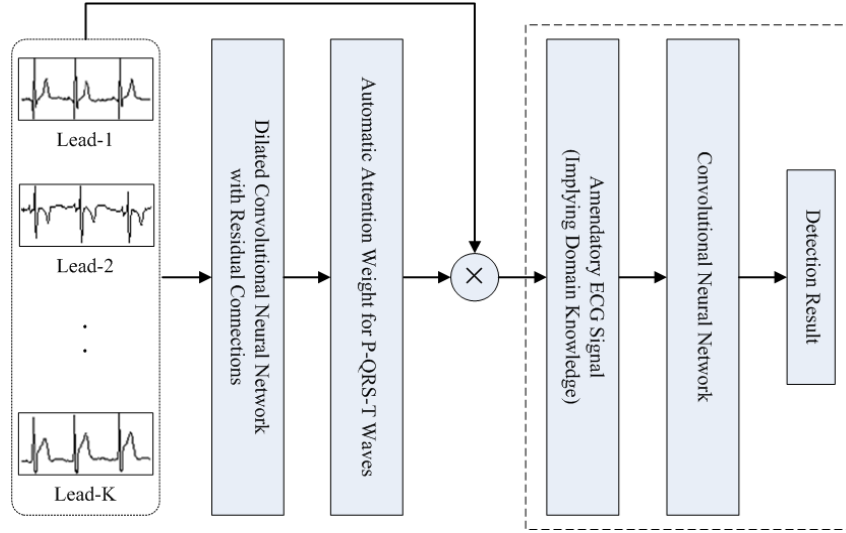

Fig.3 Example for the illustration of the proposed DANet

For the waveform enhanced module, the output size is always $1 \times frameC$ no matter whether *channelC* is greater than 1 if a *channelC* $\times$ *frameC* ECG data is input. There are four strategies to construct a dilated convolutional neural network with residual connection: (1) Only a single-lead ECG signal is selected (e.g. lead II) to input into a network model; (2) We feed each of *channelC* ECG signals into a shared network model and take an average of automatic attention weights; (3) Each of *channelC* ECG signals is input into an independent network model and their outputs are averaged; (4) We directly feed an all-lead ECG signal into a network model. Which strategy we should choose in practical use depends on the balance between classification performance and computational burdens.

In addition, we can also introduce lead attention module [7] into the classification module to strengthen or weaken the feature information derived from a specific ECG lead. It is somewhat different from the waveform enhanced module. On one hand, the lead attention score is a scalar (rather than a vector) that is multiplied by each sampling point of the corresponding lead, and thus this module has the characteristics of partial global scaling factors. On the other hand, it does not need to be trained in advance and just be integrated into the classification module as a regular component. An alternative method is that we first pre-train it by using manual lead attention scores (manually assigned according to the clinical diagnostic rule of a given disease type) and then fine-tune it together with the waveform enhanced and classification modules. There are two strategies to construct the lead attention module: each of *channelC* ECG signals is fed into its own or a shared network model. As for network model, classic neural networks such as multilayer perceptron and CNN are all competent for this task.

Model interpretability is a very important research topic in the healthcare applications. In our proposed DANet, automatic attention weights provided by the waveform enhanced module shows impressive interpretability results. From these weights we can know which waveform region (P wave, QRS complex or T wave) the classification module pays closer attention to. For a given disease type, it will lose the clinical significance if the waveform region most concerned by the proposed DANet does not consist with the cardiologists while interpreting ECG, although the final detection result may be correct. Thus, the waveform enhanced module plays a crucial part in our framework architecture. Another aspect of our model interpretability is the automatic lead attention scores, which provide the ranking of lead importance. We can figure out whether the final detection is valid by comparing these scores with the ones in cardiologists' mind.

## 3.3 Hard-coding Attention of ECG waveforms

The proposed DANet is a deep learning model based on soft-coding attention of ECG waveforms in nature: beside the routine classification module, it also includes the waveform enhanced module that can extract every P wave, every QRS complex and every T wave in ECG signals and also has the adjustability characteristic. Since traditional fiducial point methods such as ECGPuWave cannot always find P-QRS-T waves accurately and sometimes even cannot find any of them, manual attention weights are not always correct. Accordingly, the pre-trained waveform enhanced module may provide incorrect and misleading information to the classification module, so we have to update it in the later training stage to fix this problem to some extent.

However, we also find that a routine deep learning classifier can achieve better performance when taking amendatory ECG signals obtained using manual attention weights as inputs. This hard-cording attention of ECG waveform based model (called DANet-h) does not contain a feature extraction-oriented deep learning model, and thus it has much lower computational burdens compared with the DANet. The bad news is that the external software package such as ECGPuWave is needed and manual attention weights do not take advantage of data-driven technologies. Fig.3 shows the internal schematic diagram of the proposed DANet-h. Likewise, the convolutional neural network served as the classification module can be replaced by any existing deep learning model, thereby processing fixed-length or variable-length ECG signals.

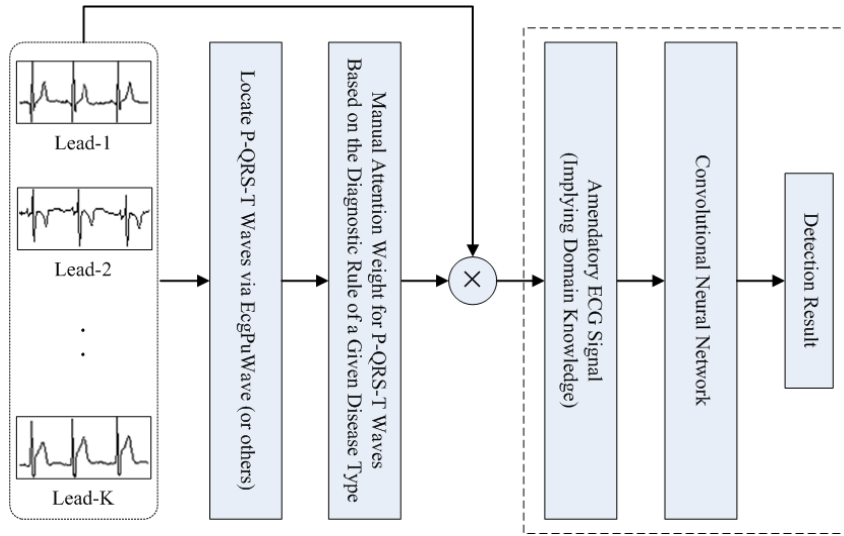

Fig.4 Example for the illustration of a DANet-h

# 4. Case Study

In this section, we will apply the proposed DANet and DANet-h models to the APC detection task using single-lead short ECG recordings. We will present the details of experimental implementation first, and then provide the results of a comparative study.

## 4.1 Detailed Implementation

Atrial premature contraction is a sign for disturbance in the depolarization process preceding the appearance of atrial fibrillation and supraventricular tachycardia in many cases, and thus timely and accurate detection of APC is of great significance in clinical practice [31]. Many scholars have carried out researches for this topic and the possible methods include feature engineering-based and deep learning-based approaches. In this paper, we take the APC detection task as a case study to verify the effectiveness of our proposed DANet.

The Aliyun-Tianchi ECG Challenge Database 2019 (AliyunDB2019) [32] provides 32,142 eight-lead ECG recordings, of which 24,106 have rhythm annotations. Each ECG recording is 10 seconds in duration and digitized at 500 samples per second. Since the 8,036 recordings do not have data annotation, they are removed from the numerical experiment. The training dataset contains 689 APC recordings and 10174 Non-APC recordings, while the testing dataset consists of 762 APC recordings and 11284 Non-APC recordings. As for the remaining 75 APC recordings and 1122 Non-APC recordings, they are collected to form the validation dataset. Note that APC here denotes atrial premature contraction, non-conducted atrial premature contraction and superventricular premature contraction while Non-APC includes all other rhythms types.

We first downsampled a raw ECG signal from 500Hz to 150Hz, and then filtered it with a 6-order bandpass filter with a cut-off frequency of 0.5Hz-50Hz [33]. To lessen the computational burden, we only selected lead II and thus the size of the resulting ECG signal is $1 \times 1500$. At the same time, the raw ECG signal was fed into the open-source software package ECGPuWave to find the onsets and offset of every P wave, every QRS complex and every T wave. The attention weight was set to 1 if a sampling point is located in P-wave regions, otherwise it was 0.3. Fig.5 shows a preprocessed ECG signal with the associated manual/automatic attention weights. Note that fiducial points found by ECGPuWave do not always appear at right locations.

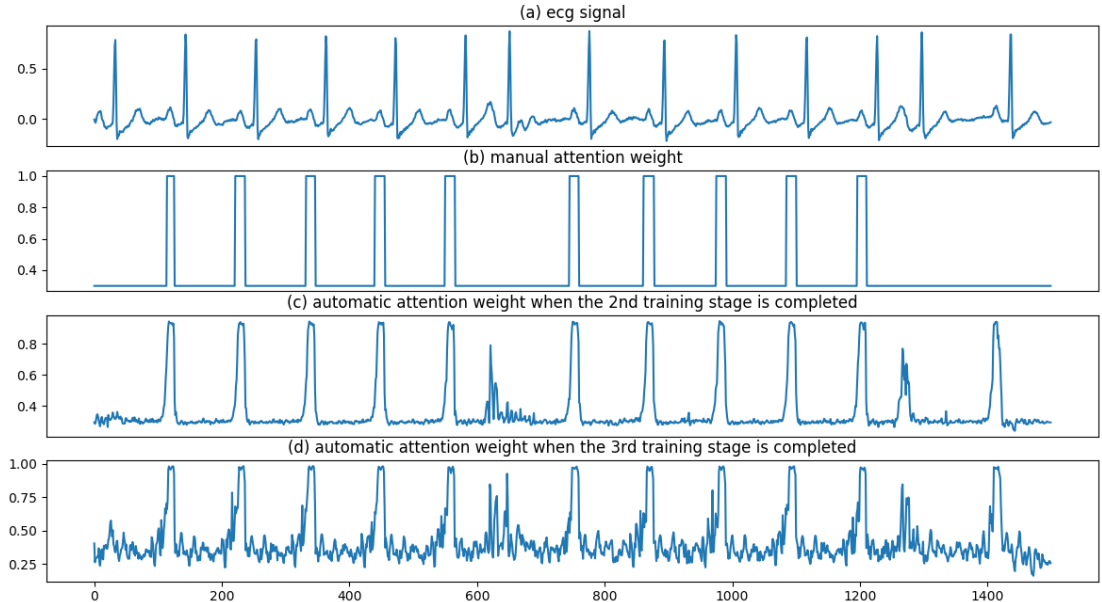

Fig.5 The illustration of an ECG signal with associated attention weights

We used a 9-layer classical convolutional neural network as the classification module in both the DANet and DANet-h models, which contains three alternating layers of convolution and max pooling, a multilayer perceptron composed of a fully-connected layer with 50 nodes and a logistic regression layer. The three kernel sizes are (21,7), (13,6) and (9,6) in turn, and the numbers of convolutional filters are set to 6, 7 and 5 respectively. For the DANet model, we also employed an 8-layer dilated convolutional neural network with residual connections [34] as the waveform enhanced module, which is composed of four dilated convolutional layers with 6 filters, a kernel size of 9 and a dilated rate of 7, and one convolutional layer with 1 filters and a kernel size of 1. The residual connection is used between the left and right branches of each even dilated convolutional layer (e.g., the 2nd and 4th layers).

Both the DANet and DANet-h models were trained in an end-to-end fashion by using adaptive moment estimation (Adam) optimizer [35]. The hyper-parameters such as learning rate and smoothing parameters for the Adam optimizer are left at their default values in Keras [36] and the mini-batch size is set to 256. For the DANet-h model, we used binary cross entropy as the loss function and trained it for 100 epochs. For the DANet model, we adopted mean squared error and binary cross entropy as the loss function for the waveform enhanced and classification modules respectively, and used 100 epochs for each of three training stages. To eliminate the influence of potential factors on performance, none of the data augmentation methods were adopted in the experiment. We also employed the standalone classification module mentioned above (namely the 9-layer CNN) for APC detection. To make a fair comparison, the training strategy and all relevant hyper-parameters remain unchanged. The only difference is that the CNN accepts original ECG recordings rather than amendatory ECG signals implying domain knowledge.

## 4.2 Results

We used four metrics including sensitivity ($Se$), specificity ($Sp$), accuracy ($Acc$) and $F$-score ($F_{APC}$, $F_{Non-APC}$ and $F_{AVG}$) to evaluate model performance. Table 1 shows a comparison of the experimental results and we highlighted the best result for each metrics in boldface.

Table 1 Detection performance comparisons of APC versus other rhythms

| Model | $Se$ | $Sp$ | $Acc$ | $F_{APC}$ | $F_{Non-APC}$ | $F_{AVG}$ |
|---|---|---|---|---|---|---|
| CNN | **56.17%** | 89.63% | 87.51% | 36.27% | 93.08% | 64.68% |
| stage-2 DANet | 54.33% | 91.50% | 89.15% | 38.78% | 94.05% | 66.41% |
| stage-3 DANet | 45.54% | **94.84%** | **91.72%** | **41.04%** | **95.55%** | **68.30%** |
| DANet-h | 55.77% | 91.14% | 88.90% | 38.87% | 93.90% | 66.38% |

From the comparison results, we can observe that both the DANet and DANet-h outperform the CNN (the baseline model) in all the metrics except sensitivity. Specifically, the stage-2 DANet (the one obtained in the second training stage) improves the performance for APC detection and reaches 1.87% increment in specificity, 1.64% increment in accuracy, 1.73% increment in $F$-score. From Table 1 we also find that the stage-2 DANet and DANet-h are evenly matched. For the stage-3 DANet model, it further increases the performance metrics and achieves an accuracy of 91.72% and an $F$-score of 68.30%. In addition, the proposed disease-specific attention mechanism also provides us the details involved in the inference process of deep learning models. As shown in Fig.5, the P-wave regions in the ECG signal get more attention for the APC detection task. Besides, we also find that the soft-coding waveform enhanced module has more powerful in locating fiducial points compared with the existing software package ECGPuWave.

## 5. Discussion

Deep learning has achieved great success in the field of ECG analysis, but they often lack model interpretability that is very important in the healthcare application. As a result, we cannot get any valuable reference information except detection results. Moreover, due to the characteristics of ECG signals (i.e. QRS complexes have more energy and more prominent features than P waves and T waves), routine deep learning models mainly capture discriminated features derived from QRS complexes. It is very unreasonable, especially when detecting disease types associated principally with non-QRS complexes from ECG signals. In our previous research work, we found that by using the same deep-learning model and the same training samples, the detection accuracy of ventricular premature contraction or atrial fibrillation is much higher than that of atrial premature contraction or ST-T abnormalities. The top-1 model wining the 2019 Tianchi Hefei High-Tech Cup ECG Human-Machine Intelligence Competition also reported similar experimental results [37]. With the aforementioned core point in mind, we have following results of a thorough analysis: (1) compared with APC, there is no need to pay special attention to the P wave for ventricular premature contraction; (2) although the diagnostic rule of "the fibrillation wave or rapid oscillations replace the P wave" should be taken into account [38], we can accurately detect atrial fibrillation from ECG signals in most cases only via the rule of "RR intervals are irregular" [39], which happen to be achieved by capturing discriminative features from the QRS complex; (3) as for ventricular premature contraction, the case is similar to that of atrial fibrillation.

The attention mechanism has the ability to make deep learning models treat different waveform regions of an ECG signal differently, but existing schemes often throw away domain knowledge and only utilize the data driven technology to obtain attention weights. Although data-driven models such as deep learning are good at extracting statistical features from input data, statistical features are not necessarily clinical features of ECG waveforms. Due to this reason, there is not always a focus on the expected waveform region for a given disease type. In this case, attention mechanism loses the significance of clinical interpretability and the diagnosis is made based on other unknown statistical features. This is obviously not the computer-aided ECG diagnostic tool physicians want in a real clinical setting.

Our proposed DANet deals with the above-mentioned problem in a different manner. To mimics the physician's diagnostic process in ECG interpretation, a specialized attention mechanism with the guidance of cardiological domain knowledge is introduced into existing deep learning models. As an implementation of this scheme, we transfer knowledge acquired in ECGPuWave to a waveform enhanced module through self-supervised pre-training and associate attention weights for P-QRS-T waves with rules for diagnosis of cardiac diseases. To compensate for defects induced by ECGPuWave, we fuse the waveform enhanced and classification modules into a single learning body and perform fine-tuning in the final training stage. Besides, the hard-coding attention mechanism of ECG waveforms that is completely independent of the end-to-end learning framework can also handle the problem of model interpretability. In this solution, waveform regions of an ECG signal followed with interest for a given disease type are always the expected ones (there may be problems with precision at most) except those in which P-QRS-T waves cannot be located by ECGPuWave. Thus, the case in which detection results given by deep neural networks are correct while the most discriminated features are extracted from task-unconcerned waveform regions can be avoided. In brief, the novelty of the proposed DANet is that the diagnostic rule for a given disease type is represented by attention weights in an explicit manner, rather than in an ambiguous manner. This will greatly enhance the effectiveness and robustness of deep learning models.

# 6. Conclusion

In this study, we develop a medical explainable deep learning model named DANet for arrhythmia detection from short ECG recordings. It utilizes a disease-specific attention mechanism to amend original ECG signals before feeding them into the classification module, thus cardiological domain knowledge can be taken into account when deep learning models make predictions. Using a case study with respect to APC detection, the effectiveness and potential of our model is validated. The proposed DANet (or DANet-h) produces better performance in detecting APC on the AliyunDB2019 compared with the benchmark model. We also have carries out experiments and found that it still can improve detection performance of ST-T abnormalities [33]. Besides, the proposed DANet also provides a degree of model interpretability: the attention weights of ECG waveforms implying domain knowledge are always put together with a detection result, which is a very important requirement in the healthcare applications. In the future, we will extend the proposed model to detect other arrhythmia types from ECG signals.

# Acknowledgement

This research was supported by Zhejiang Provincial Natural Science Foundation of China under Grant No.LQ20F020020 and Scientific Research Start-up Foundation of Hangzhou Normal University.